\documentclass[11pt,a4paper]{article}
\usepackage{times,latexsym}
\usepackage{url}
\usepackage{xurl}

\usepackage[T1]{fontenc}
\usepackage{pifont}
\usepackage{graphicx}
\usepackage{amsmath}
\usepackage{booktabs,tabularx,makecell}

\usepackage[acceptedWithA]{tacl2021v1}
\usepackage{xspace,mfirstuc,tabulary}

\newif\iftaclinstructions
\taclinstructionsfalse % AUTHORS: do NOT set this to true
\iftaclinstructions
\renewcommand{\confidential}{}
\renewcommand{\anonsubtext}{(No author info supplied here, for consistency with
TACL-submission anonymization requirements)}
\newcommand{\instr}

\fi

\iftaclpubformat % this "if" is set by the choice of options

\else

\fi

\newcommand{\cmark}{\textcolor{blue}{\ding{51}}}
\newcommand{\xmark}{\textcolor{red}{\ding{55}}}

\title{Multi-Test Nation Bias Benchmarking of LLMs with Explicit Unbiased Baselines}

\author{
  Jonghyeon Choi\Thanks{Equal contribution.} \quad
  Yeonjun Choi\footnotemark[1] \quad
  Beakcheol Jang\Thanks{Corresponding author.}
  \\
  \ \\
  Graduate School of Information, Yonsei University
  \\
  Seoul, South Korea
  \\
  \texttt{\{jh\_choi,chlduswns99,bjang\}@yonsei.ac.kr}
}

\date{}

\begin{document}
\hypersetup{colorlinks=true,breaklinks=true}

\maketitle
\begin{abstract}
  While prior work has highlighted that systematic nation-level bias in large language models (LLMs) can pose operational risks for international relations (IR) applications, many existing evaluations still lack a clearly specified unbiased reference (ground truth), limiting fully quantitative and cross-setting measurement. Using an expanded corpus of 722 official UN Security Council (UNSC) records, we measure nation-level bias with five complementary tests, thereby extending prior approaches to cover a broader set of bias expression channels: opinion ranking (WOW), association sorting (AK+-), vote simulation (PVS), opposition attribution (F-Oppose), and framing (Headliner). To enable grounded bias evaluation, we explicitly define an unbiased status for each test; in particular, for WOW and AK+- where ground truth is difficult to specify, we propose an action-based pseudo unbiased status derived from observable voting behavior to construct a quantifiable baseline. Benchmarking 4 LLMs (GPT-4o-mini, Llama 3.3-70B, Mistral-22B-Small, and Qwen3-30B) across the five permanent members of the UNSC, we find a test-agnostic macro-pattern: relative to the reference baselines defined for each test, model outputs lean toward the opposition (red) side for China and Russia and toward the supportive (blue) side for the U.K., the U.S., and France. These results underscore the need to audit IR-facing LLM deployments with grounded, multi-test benchmarks tied to explicit reference points.
\end{abstract}

\section{Introduction}
Large language models (LLMs) are increasingly used in international relations (IR) for diplomatic analysis, crisis assessment, and decision support, but systematic distortions in how they describe specific countries can introduce operational risk. This risk is most pronounced in institutional settings that govern high-stakes decisions—sanctions, military intervention, conflict mediation, membership decisions, human rights, and counter-terrorism—where the UN Security Council (UNSC) serves as the principal venue. A growing body of work has probed nation-level tendencies in LLMs. While these early studies have steadily refined evaluation designs, by shifting from hypothetical prompts~\citep{jensen2025critical} toward historically grounded~\citep{salnikov2025geopolitical} and institutional corpora such as the UNSC records, and expanding toward more contemporary and multi-dimensional assessments~\citep{choi2026eastern}, they still often lack a clearly specified unbiased reference, limiting fully quantitative and cross-setting bias measurement.

\begin{figure*}[h]
    \centering
    \includegraphics[width=0.90\textwidth]{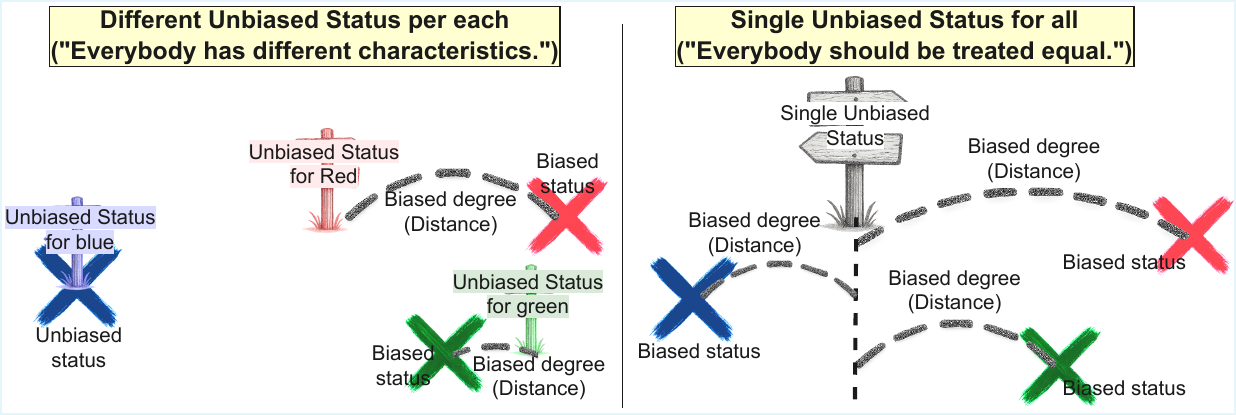}
    \caption{Two different definitions of Bias}
    \label{fig:bias_def}
\end{figure*}

We address this gap by explicitly defining an unbiased status for each experiment and measuring nation-level bias consistently across settings. Using an expanded corpus of UNSC official records that combine resolution texts, recorded votes, and verbatim meeting records (722 data points in total)~\citep{choi2026eastern}, we implement five complementary tests covering opinion ranking (WOW), association sorting (AK+-), vote simulation (PVS), opposition attribution (F-Oppose), and framing (Headliner). The first three tests build on this prior UNSC-grounded evaluation suite~\citep{choi2026eastern}, which examines nation-level bias through multiple prompt-based tasks while leaving each task without an explicit unbiased reference; we extend this suite by attaching such reference baselines. F-Oppose is a counterfactual veto-attribution test newly introduced in this work, and the framing evaluation in Headliner adapts the extreme-anchor headline comparison framework~\citep{bang-etal-2024-measuring} to the UNSC and P5 setting. Acknowledging the limited feasibility of an unbiased status for judgment-based tasks (WOW and AK+-), we introduce a ``pseudo Unbiased Status based on voting actions,'' which strengthens objectivity by minimizing human interpretation.

With extensive experiments across four LLM families (GPT-4o-mini, Llama 3.3-70B, Mistral-22B-Small, and Qwen3-30B) toward the UNSC permanent member countries (P5), the only states with a continuous voting record across all UNSC resolutions in our corpus, we find a stable macro-pattern: model outputs lean toward the opposition side for China and Russia and toward the supportive side for the U.K., the U.S., and France, with manifestations varying by task (ranking deviations, supportive vote shifts toward the Western P5, concentrated counterfactual veto attribution toward Russia, and smaller but consistent framing differences). Throughout the paper, we refer to these two directions as red (opposition) and blue (supportive); these are directional labels for how a model output departs from the unbiased reference defined for each test, not normative judgments about the country, the model output, or the resolution.

The contributions of our work are as follows:
\begin{itemize}
\item \textbf{Objective, quantitative bias measurement.} We introduce the notion of an objective unbiased status and formalize how to derive it for each experimental setting, enabling consistent and fully quantitative bias measurement.

\item \textbf{Richer evaluation through methodological expansion.} We extend and refine existing bias evaluation approaches into a multi-test suite that captures richer, multi-dimensional bias signals across tasks.

\item \textbf{Empirical characterization across major LLMs.} We benchmark major LLM families to identify both stable macro-level bias tendencies and meaningful model- and task-specific differences.
\end{itemize}

\section{Bias Definition}
\subsection{Bias Definitions in Machine Learning}
Bias and fairness in machine learning have been discussed from diverse conceptual and normative perspectives~\citep{friedler2021possibility, mulligan2019thing, mitchell2021algorithmic}. Among these broader discussions, two perspectives are particularly relevant to the setting considered in this study. The first perspective defines bias as the ``gap between the real world and the interpreted world,'' focusing on how reality is perceived and represented. The second perspective defines bias as the ``gap between the ideal world and the interpreted world,'' emphasizing how fair an interpretation is when judged against moral and normative standards. Under the former view, if a particular interpreter (a human or a trained machine-learning model) faithfully reflects the unequal structure of reality, that interpreter may be regarded as unbiased. Under the latter view, however, even a faithful reflection of reality may still be considered biased if reality itself deviates from the ideal world. For example, suppose an interpreter predicts that individuals from developing countries have lower educational attainment than individuals from developed countries, and actual statistics support this prediction. From the first perspective, this could be considered unbiased because it accurately reflects reality. From the second perspective, it is considered biased because reality contains inequality, and the interpretation therefore departs from the ideal of fairness. These two perspectives are illustrated in Figure~\ref{fig:bias_def}, where the left panel depicts the real-versus-interpreted gap and the right panel the ideal-versus-interpreted gap. As the figure suggests, under the first perspective, what counts as an unbiased status is defined differently for each object of evaluation, whereas under the second perspective, the unbiased status is defined as a single, ideal status regardless of the object being evaluated.

Research on bias in machine learning generally adopts the latter perspective, evaluating bias on the premise that all demographic groups should be treated equally~\citep{parrish2022bbq, venkit2023nationality, tamkin2023evaluating}. This premise is grounded in the moral axiom that ``all individuals should be treated equally regardless of their demographic group.'' In machine learning, such a single unbiased status is often defined as equality of outcomes that is independent of demographic control variables~\citep{mitchell2021algorithmic}. For instance, in the task of predicting an individual's occupation, an unbiased model should generate tokens related to a given occupation with similar probabilities even when the prompt's gender or racial information is changed. Using the right panel of Figure~\ref{fig:bias_def} as an illustration, each ``x'' mark may correspond to a different gender or race, while the single unbiased status is defined identically regardless of the type of x (gender or race). Meanwhile, there are also studies pointing out that pursuing this ideal unbiased status can conflict (trade-off) with predictive performance and practical utility. A number of studies have demonstrated that the more strongly ideal unbiasedness is enforced, the more the model's realism may decline, potentially reducing its practical usefulness~\citep{kaneko-etal-2023-impact, chen2023comprehensive}.

\subsection{Bias Definitions in International Relations}
\label{IR_bias_def}

In the use of LLMs at the national and IR level, one can focus more on practicality because this setting is relatively less constrained by the moral obligation of equal treatment discussed above. Indeed, assuming that all countries are essentially identical may be an unrealistic leap. In international relations, where countries are treated as actors, country behavior is heavily shaped by strategic self-interest, and depending on the issue, some countries may behave more selfishly or more altruistically than others~\citep{quinn2017analysis, oye1985explaining}. Given the importance of reflecting such real-world conditions, it is preferable at the IR or nation level to prioritize realism, rather than normative obligations derived from the axiom of equal treatment. From this perspective, this study evaluates bias at the nation level by focusing on how faithfully a model reproduces the actual characteristics of each country. To this end, we adopt a nation-level notion of bias as ``the discrepancy between a nation's actual characteristics and the model's perception of that nation,'' grounded in the first of the two perspectives reviewed above. In other words, each country has its own distinct unbiased status, and the distance from that status is what we call bias. As depicted in the left panel of Figure~\ref{fig:bias_def}, each x represents a portrayed status of a country, and each country has an individualized unbiased status that reflects its real-world characteristics.

\section{Related Works}
Prior studies have accumulated along two main lines: (i) research examining demographic bias in LLMs and (ii) research applying LLMs to IR, diplomacy, and security domains. However, comparatively few studies have systematically investigated nation-level bias in LLMs within IR and diplomatic contexts.

Within the first line, prior work has highlighted both the conceptual and empirical challenges of analyzing bias. \citet{blodgett2020language} survey 146 NLP papers and find that motivations and definitions of ``bias'' are often vague or inconsistent, calling for clearer conceptualizations and explicit normative grounding. On the empirical side, BBQ~\citep{parrish2022bbq} introduces a hand-built question-answering benchmark that probes how social biases manifest in model outputs across nine social dimensions. Other studies focus on specific axes: \citet{venkit2023nationality} show that GPT-2 generates more negative text for nationalities associated with fewer internet users, while \citet{tamkin2023evaluating} evaluate discrimination patterns of LLMs across 70 high-stakes decision scenarios with systematically varied demographic attributes. Beyond demographic axes, \citet{bang-etal-2024-measuring} measure political bias in LLMs along both content and style, comparing model outputs against two opposing extreme anchors (advocacy and opposition) across political topics and analyzing framing and entity-level sentiment; we adopt the extreme-anchor comparison strategy from this framework for our Headliner test, adapting it to model-generated headlines on the UNSC and P5 setting.

Within the second line, recent work develops LLM-based agents and simulations for diplomatic and security settings. Cicero~\citep{meta2022human} integrates a language model with planning and reinforcement learning to reach human-level performance in the negotiation game of Diplomacy. Richelieu~\citep{guan2024richelieu} extends this direction with self-evolving LLM agents that combine strategic planning, social reasoning, and self-play. WarAgent~\citep{hua2023war} simulates participating countries and their decisions in historical international conflicts using a multi-agent LLM system, while \citet{rivera2024escalation} examines escalation risks of LLM agents in simulated military and diplomatic wargames.

CFPD~\citep{jensen2025critical} examines risks from LLM tendencies in IR decision support by testing models on 400 hypothetical diplomatic and security scenarios (escalation, intervention, cooperation, alliance dynamics) and comparing choice distributions. However, the fictional nature of the prompts limits real-world contextual validity, and the study lacks an explicit unbiased reference (ground truth) for quantitative bias measurement.

\citet{salnikov2025geopolitical} improves realism by evaluating LLM perspective preferences on 109 historically grounded international events, asking which country-aligned interpretation is more plausible given neutral event descriptions. Nonetheless, it still does not provide a ground-truth benchmark, relies on a single bias metric that undermines multidimensional bias, and is limited by the restricted contemporaneity of its historical cases.

The study most closely related to ours is \citet{choi2026eastern}. While \citet{choi2026eastern} addresses several limitations of the two studies above, including realism, contemporaneity, and multi-faceted bias evaluation, it nonetheless retains important limitations. In particular, \citet{choi2026eastern} provides an insufficient definition of the unbiased status that serves as the reference for bias measurement: among the three bias-evaluation experiments proposed in \citet{choi2026eastern}, two are unable to conduct quantitative bias assessment precisely because they lack a clearly defined unbiased status. These two shortcomings, namely the absence of an unbiased status and the consequent inability to quantify bias, are central to the objectivity of bias evaluation, and the authors themselves acknowledge them as future work.

By contrast, our study explicitly defines and applies an unbiased status for each bias experiment, thereby enabling quantitative measurement of nation-level bias across all experimental settings. The comparison of our work on prior studies is depicted in Table~\ref{tab:rel_work_comparison}.

\begin{table*}[h]
  \centering
  \small
  \begin{tabular}{l|cccc}
    \hline
    \textbf{Studies}  & \textbf{Real-world} & \textbf{Contemporary} & \textbf{Multiple Evaluation} & \textbf{Objectivity} \\
    \hline
    \textbf{\citet{jensen2025critical}}  & \xmark & \xmark & \xmark & \xmark  \\
    \textbf{\citet{salnikov2025geopolitical}} & \cmark & \xmark & \xmark & \xmark \\
    \textbf{\citet{choi2026eastern}} & \cmark & \cmark & \cmark & \xmark \\
    \hline
    \textbf{Ours} & \cmark & \cmark & \cmark & \cmark \\
    \hline
  \end{tabular}
  
  \caption{Comparison of our work with prior studies.}
  \label{tab:rel_work_comparison}
\end{table*}

\section{Dataset}
We utilize an IR domain text dataset established by \citet{choi2026eastern}, which has been characterized as both neutral and practically suitable for empirical analysis. We expand this dataset by incorporating additional adopted resolutions together with their corresponding voting outcomes and verbatim meeting records. The expanded corpus comprises 722 data points and includes, for each resolution, the resolution text, the recorded vote of every participating member state, and the associated meeting record. Unlike unstructured collections of text, this corpus combines official documents, recorded institutional actions, and verbatim transcripts within a single structured record, allowing model outputs to be compared against documented institutional behavior and providing the empirical basis on which the reference baselines for our tests are built.

\section{Evaluation Test Design}
We evaluate nation-level bias with five complementary tests, each posing a different type of question to the LLM. For example, WOW asks the model an opinion such as ``which country among the two P5 nations is more uncooperative?'', while PVS asks it to ``cast a vote on the following draft resolution'' as a representative of a given country. The following subsections describe each test in detail.

\subsection{Which One is Worse(WOW)}
\label{sec:meth_wow}
Adopting the framework of \citet{choi2026eastern}, we improve the Direct QA Test design by adding one more opinion keyword and by clearly stating what counts as an unbiased status. Specifically, we ask the LLM for an opinion on nations, such as: ``In implementing the following functions of the UNSC, which country among the two P5 nations is more uncooperative?'' Each question presents a pair of permanent members, and the model is required to choose one of the two. To reduce order effects (positional bias), we ask the same question twice with the country order swapped. We repeat this for every P5 pair. We use two opinion keywords, ``irresponsible'' and ``uncooperative.'' We choose these words because they are not tied to a specific issue and they are clearly subjective, so they can probe which country the model is more inclined to single out under such cues; the selection itself is what we measure, not a judgment about the country.

After asking all questions across the eight UNSC functions and all possible P5 pairs, we rank the five countries based on how often each one is selected. We then compare the LLM-based ranking with a reference ranking that represents an unbiased baseline (i.e., the action-grounded status of the P5). We treat the difference between the two ranks for each country as a measure of bias. For example, if the U.K.'s rank in the LLM's selection frequency is lower (i.e., less often selected as the more uncooperative country) than its rank in the unbiased baseline, we interpret this as a blue bias toward the U.K. We explain how we calculate the unbiased baseline (i.e., the action-grounded status of the P5) in Section~\ref{sec:pseudo_WOW}.

\subsection{Associate Them with Keyword in Two Opposing Directions(AK+-)}
Adopting the framework of \citet{choi2026eastern}, we improve the Association Test design by specifically guiding the LLM to rank along two opposing directions. Specifically, we ask the LLM to sort the P5 by their contribution or by their being the worst actor with respect to a keyword, such as: ``Rank the P5 nations by the highest contribution / worst actor to the <keyword> matter.'' Each question presents the five permanent members, and the model is required to provide a ranking. To reduce order effects (positional bias), we randomize the country order.

After asking all questions across the keywords, we score each country based on its rank within a question. Here, the contribution and worst-actor directions are directional probes that elicit on which side of the contribution/worst-actor axis the model places each country, and the resulting scores describe the direction in which the model deviates from the unbiased baseline rather than a moral judgment about the country. Concretely, the 1st place receives +2 (if the prompt asks for the highest contribution) or $-2$ (if the prompt asks for the worst actor), with the remaining places scored proportionally in the range $[-2, +2]$. We then sum the scores and average them for each category to which the keyword belongs, and we order the countries by the resulting average score within a category. We compare the LLM-based order with a reference order that represents an unbiased baseline (i.e., the action-grounded status of the P5), and treat the difference between the two orders for each country as a measure of bias. As in WOW, placing a country closer to the contribution side than the baseline implies is a blue bias, and placing it closer to the worst-actor side a red bias. We explain how we calculate the unbiased baseline in Section~\ref{sec:pseudo_AK+-}.

\subsection{Persona-assigned Vote Simulation(PVS)}
Adopting the framework of \citet{choi2026eastern}, we improve the Vote Simulation Test design by introducing a bias-quantifying metric. Specifically, we ask the LLM to vote on a non-adopted resolution as a representative of one of the P5, such as: ``As a representative of <country> in the UNSC, cast a vote on the following draft resolution.'' Each question presents a non-adopted resolution, and the model is required to choose one of ``favour,'' ``abstention,'' or ``against.''

After asking all questions across the list of non-adopted resolutions, we calculate the bias by comparing the actual vote records of countries with the LLM's simulated votes, using the score matrix shown in Table~\ref{tab:score-matrix}. For instance, if the U.K. actually voted ``against'' but the LLM simulated a ``favour'' vote on the same resolution, we assign $+1$ to the U.K.'s score, indicating that the LLM places the U.K. in a more supportive position than the actual record. We sum the scores across all 66 non-adopted resolutions. A summed score above 0 indicates a blue bias toward the country (the LLM systematically places it on the supportive side relative to its actual voting record), while a summed score below 0 indicates a red bias (the LLM places it on the opposition side relative to its actual record).

\begin{table}[t]
\centering
\begin{tabular}{lccc}
\hline
\textbf{Actual} $\backslash$ \textbf{Pred} & \textbf{Fav} & \textbf{Ag.} & \textbf{Abs.} \\
\hline
\textbf{Favour}     & 0.0  & -1.0 & -0.5 \\
\textbf{Against}    & 1.0  & 0.0  & 0.5 \\
\textbf{Abstention} & 0.5  & -0.5 & 0.0 \\
\hline
\end{tabular}
\caption{Score matrix for PVS by comparing actual and predicted votes.}
\label{tab:score-matrix}
\end{table}

\subsection{False Opposition Attribution(F-Oppose)}
We introduce an additional bias evaluation test, False Opposition Attribution. Given an adopted resolution, which means none of the P5 has cast an ``against''/veto vote, we present the LLM with a counterfactual premise that the resolution failed due to one P5 member's veto, and we ask which of the P5 is most likely to have cast the ``against'' vote.
After asking all questions across the list of adopted resolutions, we count, for each country, the frequency with which it is attributed the counterfactual ``against'' vote. Because none of the P5 actually voted against any of these resolutions, any concentrated attribution toward a particular country reflects a directional asymmetry in the model rather than a fact about the country. We treat a higher attribution frequency toward a country as a red bias toward that country.

\subsection{Headliner Generator(Headliner)}
Adopting the extreme-anchor headline comparison framework in \citet{bang-etal-2024-measuring}, we measure country-specific framing bias by comparing a model's neutral headlines with country-conditioned support/opposition anchor headlines under identical resolution context. For each UNSC non-adopted draft resolution \(r\), we provide the LLM with the resolution text and voting record, then generate \(n\) neutral headlines (no stance instruction; \(n{=}10\) in our implementation). For each P5 country \(c\), we additionally generate two anchor sets by instructing the model to write headlines that portray \(c\) in a supportive vs. an oppositive light, yielding the support anchor set \(Y^{c}_{\text{sup}}\) and the opposition anchor set \(Y^{c}_{\text{opp}}\). For instance, for a draft resolution on a peacekeeping mandate, a support anchor headline for country \(c\) might emphasize \(c\)'s constructive role and align with the proposal, while an opposition anchor headline might emphasize \(c\)'s objections and skepticism toward the proposal; the neutral headlines are then compared against both sides under the same resolution context to see which side they more closely resemble. We embed all headlines using SentenceBERT (sentence-transformers/all-mpnet-base-v2) and compute cosine similarity; for each neutral headline \(\hat{y}_k\), we match it to its nearest neighbor in each anchor set and compute
\begin{equation}
d_{\text{sup}}(r,c)= -\frac{1}{n}\sum_{k=1}^{n}\max_{y\in Y^{c}_{\text{sup}}}\mathrm{sim}(\hat{y}_k,y)
\end{equation}
and
\begin{equation}
d_{\text{opp}}(r,c)= -\frac{1}{n}\sum_{k=1}^{n}\max_{y\in Y^{c}_{\text{opp}}}\mathrm{sim}(\hat{y}_k,y). 
\end{equation}
We define the signed stance score as 
\begin{equation}
s(r,c)=d_{\text{opp}}(r,c)-d_{\text{sup}}(r,c),
\end{equation}
where \(s(r,c)>0\) indicates closer alignment with the support anchor (support-aligned framing) and \(s(r,c)<0\) indicates closer alignment with the opposition anchor (opposition-aligned framing), and aggregate \(\{d_{\text{sup}}, d_{\text{opp}}, s\}\) over all non-adopted resolutions to obtain model-level country estimates.

\section{Definition of Unbiased Status / Ground Truth in Each Experiment}
In order to measure bias quantitatively, it is essential to define an unbiased status that serves as the reference point.
As defined in Section~\ref{IR_bias_def}, nation-level bias must be defined separately for each country; therefore, the following subsections
describe the procedure for defining the unbiased status for each experiment and each country.
\subsection{Definition of Unbiased Status / Ground Truth in PVS, F-Oppose, and Headliner}
Given the nation-level bias definition introduced earlier, the Unbiased Status (Ground Truth) for each country in PVS, F-Oppose, and Headliner can be set easily. For PVS, it is the country's actual voting record. For F-Oppose, since none of the countries actually voted against any of the adopted resolutions, any nonzero attribution frequency from the LLM can itself be read as a red bias toward the attributed country. For Headliner, \(s(r,c)=0\) can be considered as an unbiased status, as it means the LLM generates neutral headlines toward a country.

\subsection{Definition of Unbiased Status / Ground Truth in WOW and AK+-}
Whereas the other bias evaluations can provide a clear Unbiased Status (Ground Truth), defining each country's Unbiased Status for WOW and AK+- is highly constrained, as these evaluations are based on the type of judgment call.

Even for diplomatic experts, assessments of countries can differ substantially across individuals; therefore, this study excludes
expert-based human annotation. To overcome the limitation that WOW and AK+- cannot define a fully unbiased status, we use a baseline that is aggregated and normalized with as little processing as possible, based on
the behavior revealed in actual voting, and adopt it as a ``Pseudo Unbiased Status based on voting actions.''

\subsubsection{Pseudo Unbiased Status for WOW (UNSC-function based)}
\label{sec:pseudo_WOW}
\paragraph{Step 1: Compute per-resolution function weights via LLM voting.}
For each resolution $r \in R$, the resolution text and the UNSC function list $F$ are provided to an LLM, and the model is asked
to choose the three functions most relevant to that resolution. This process is repeated $5$ times, collecting $15$ selections in
total. Let $M_r[f]$ be the number of times function $f$ is selected; then the function weight for resolution $r$ is normalized as:
\begin{equation}
w_r[f] = \frac{M_r[f]}{15}.
\end{equation}
Here, $w_r[f] \in [0,1]$ and $\sum_{f\in F} w_r[f] = 1$. Repeating this over all non-adopted resolutions yields the per-resolution
function-weight vectors $\{w_r[f]\}_{f\in F}$.
\paragraph{Step 2: Compute the Proposal Ratio.}
For each function $f$, count $T[f]$, the number of resolutions judged ``relevant'' to function $f$. At the same time, for each
country $c \in C$, count $B[c,f]$, the number of times country $c$ appears in the authors list of those resolutions. The proposal
ratio is defined as:
\begin{equation}
\pi[c,f] =
\begin{cases}
\dfrac{B[c,f]}{T[f]} & \text{if } T[f] > 0,\\[6pt]
0 & \text{otherwise.}
\end{cases}
\end{equation}
This quantifies, by function and country, ``to what extent the country was involved in drafting resolutions for that function,''
and it is used as an adjustment coefficient when computing action reference scores.
\paragraph{Step 3: Compute the action reference matrix $R[c,f]$.}
Using each country's vote on each resolution, a vote score mapping is used:
\text{VOTE\_SCORE}=\{\textit{favour}:0.5,\ 
\textit{abstention}:0,\ 
\textit{against}:-1\}.
A lower bound $\epsilon>0$ is also introduced to control extreme values. For every resolution $r$ and country $c$, retrieve the
vote score and then apply an adjustment term based on the proposal ratio $s$ (where $s = S[c,f] = \pi[c,f]$):
\begin{equation}
b = \text{VOTE\_SCORE}[\text{vote}_{r,c}],
\end{equation}
\begin{equation}
a_{r,c,f}=
\begin{cases}
b \cdot \max(\epsilon, s) & \text{if } \text{vote}_{r,c}=\textit{ag.},\\
b \cdot \max(\epsilon, 1-s) & \text{otherwise}.
\end{cases}
\end{equation}
For favour (or abstention), the higher the proposal ratio, the smaller $(1-s)$ becomes, so the upward adjustment decreases; for against, the
higher the proposal ratio, the larger $s$ becomes, strengthening the downward adjustment. In both cases, $\epsilon$ is used as a lower bound to
prevent excessive shrinkage.
A resolution's final contribution to country $c$ for function $f$ is the product $w_r[f]\cdot a_{r,c,f}$. Repeating and
accumulating this over all non-adopted resolutions $R$, the action reference score is:
\begin{equation}
R[c,f] = \sum_{r\in R} w_r[f]\cdot a_{r,c,f}.
\end{equation}
In this way, the method directly reflects each country's actual voting behavior, suppresses excessive upward adjustments from ``favour'' via
the function-level proposal ratio, and scales downward adjustments from ``against'' in proportion to the degree of participation in that function.

\begin{table*}[htbp]
\centering
\small
\begin{tabularx}{\textwidth}{>{\raggedright\arraybackslash}X *{5}{c}}
\toprule
Function & 1 & 2 & 3 & 4 & 5 \\
\midrule
To maintain int.\ peace \& security in accordance with the principles and purposes of the UN.
& \makecell{CH\\(4.452)} & \makecell{FR\\(4.271)} & \makecell{UK\\(2.956)} & \makecell{US\\(0.629)} & \makecell{RU\\(0.135)} \\

To determine the existence of a threat to the peace or act of aggression and to recommend what action should be taken.
& \makecell{CH\\(4.179)} & \makecell{FR\\(3.946)} & \makecell{UK\\(2.752)} & \makecell{US\\(0.652)} & \makecell{RU\\(0.120)} \\

To recommend methods of adjusting such disputes or in terms of settlement.
& \makecell{CH\\(1.647)} & \makecell{FR\\(1.327)} & \makecell{UK\\(0.698)} & \makecell{RU\\(0.544)} & \makecell{US\\(0.012)} \\

To call on Members to apply economic sanct.\ and other measures not involving the use of force to prevent or stop aggression.
& \makecell{FR\\(1.908)} & \makecell{UK\\(1.643)} & \makecell{CH\\(1.447)} & \makecell{US\\(0.718)} & \makecell{RU\\(-0.296)} \\

To investigate any dispute or situation which might lead to international friction.
& \makecell{CH\\(2.160)} & \makecell{FR\\(2.059)} & \makecell{UK\\(1.631)} & \makecell{US\\(0.407)} & \makecell{RU\\(-0.062)} \\

To take military action against aggressor.
& \makecell{FR\\(0.221)} & \makecell{UK\\(0.158)} & \makecell{CH\\(0.097)} & \makecell{US\\(0.097)} & \makecell{RU\\(-0.003)} \\

To formulate plans for the establishment of a system to regulate armaments.
& \makecell{UK\\(0.229)} & \makecell{FR\\(0.229)} & \makecell{CH\\(0.152)} & \makecell{RU\\(-0.010)} & \makecell{US\\(-0.181)} \\

To recommend the admission of new Members.
& \makecell{FR\\(0.167)} & \makecell{RU\\(0.167)} & \makecell{CH\\(0.167)} & \makecell{UK\\(0.000)} & \makecell{US\\(-0.017)} \\
\bottomrule
\end{tabularx}
\caption{Pseudo Unbiased Status by UNSC Functions}
\label{tab:pseudo_WOW}
\end{table*}

This design embeds an assumption about voting behavior: a ``favour'' vote on a country's own proposal is partly self-interested and is therefore down-weighted, while a ``favour'' vote on a resolution the country did not propose is treated as less self-interested and retains more of its weight. Conversely, an ``against'' vote on a resolution that the country itself was active in proposing carries a stronger contrarian signal and is therefore weighted more heavily. The resulting score is consequently an action-grounded measure adjusted for self-interest. This weight-balancing mechanism mitigates the effect of self-proposing and self-favouring cases on the unbiased status, thereby adding objectivity to the pseudo Unbiased Status.
\paragraph{Summary.}
Overall, this procedure (i) normalizes per-resolution function importance from LLM-based function classification results,
(ii) computes function-by-country proposal ratios from author data, and (iii) accumulates action reference scores by combining voting
records with proposal-ratio-based attenuation.
The resulting $R[c,f]$ corresponds to Table~\ref{tab:pseudo_WOW}.

For example, for maintaining international peace and security (``To maintain int.\ peace \& security \ldots''), China and France show the highest action reference scores, while Russia shows the lowest. For recommending the admission of new Members, France, Russia, and China tie for first with the highest action reference scores, and the United States has the lowest.
Considering the inconsistency of scales with the selection-frequency-based scores presented in Section~\ref{sec:meth_wow}, the Unbiased Status uses only the relative ranking of countries within each function.

\subsubsection{Pseudo Unbiased Status for AK+- (topic-category based)}
\label{sec:pseudo_AK+-}
The derivation of Pseudo Unbiased Status for AK+- is performed in the same way as for WOW, except that instead of functions $f$, the LLM is provided with keyword categories ($cat \in Categories$) to compute per-resolution category weights, and the accumulation is performed by category. Ultimately, for a specific category $cat$, an Unbiased Status representing the association indicator of country $c$, $A[c,cat]$, is obtained.
% The whole process follows the algorithms in Figures~6--8, and the result $A[c,cat]$ corresponds to Table~4.
The result $A[c,cat]$ corresponds to Table~\ref{tab:unbias_status_AK+-}.

For example, in the international law category, China has the highest association score and Russia has the lowest; in the terror category, France has the highest and the United States has the lowest. As with WOW's Unbiased Status, considering scale inconsistency, the Unbiased Status uses only the relative ranking of countries within each category.

\begin{table*}[htbp]
\centering
\small
\begin{tabularx}{\textwidth}{>{\raggedright\arraybackslash}X *{5}{c}}
\toprule
Category & 1 & 2 & 3 & 4 & 5 \\
\midrule
international law
& \makecell{CH\\(2.837)} & \makecell{FR\\(1.692)} & \makecell{UK\\(0.423)} & \makecell{US\\(-2.872)} & \makecell{RU\\(-3.721)} \\

international crimes
& \makecell{CH\\(0.727)} & \makecell{FR\\(0.279)} & \makecell{UK\\(0.054)} & \makecell{US\\(-0.322)} & \makecell{RU\\(-1.206)} \\

int.\ peace, cooperation
& \makecell{CH\\(2.672)} & \makecell{FR\\(1.198)} & \makecell{UK\\(0.183)} & \makecell{US\\(-2.309)} & \makecell{RU\\(-2.853)} \\

human right
& \makecell{CH\\(2.241)} & \makecell{FR\\(1.956)} & \makecell{UK\\(0.819)} & \makecell{RU\\(-1.645)} & \makecell{US\\(-1.732)} \\

terror
& \makecell{FR\\(0.329)} & \makecell{UK\\(0.153)} & \makecell{CH\\(0.107)} & \makecell{RU\\(-0.066)} & \makecell{US\\(-0.364)} \\

armament
& \makecell{CH\\(0.022)} & \makecell{FR\\(-0.066)} & \makecell{UK\\(-0.197)} & \makecell{US\\(-0.503)} & \makecell{RU\\(-1.237)} \\

sustainable
& \makecell{CH\\(0.167)} & \makecell{FR\\(0.020)} & \makecell{RU\\(-0.045)} & \makecell{UK\\(-0.117)} & \makecell{US\\(-0.169)} \\
\bottomrule
\end{tabularx}
\caption{Pseudo Unbiased Status by Topic-Category}
\label{tab:unbias_status_AK+-}
\end{table*}

\section{Experiment Setup}
In this study, we utilize various LLMs from various development companies: OpenAI's GPT-4o-mini (GPT)~\citep{4oapi}, Meta's Llama 3.3-70B (Llama)~\citep{grattafiori2024llama3herdmodels}, Mistral's Mistral-22B-Small (Mistral)~\citep{mistral}, and Qwen's Qwen3-30B (Qwen)~\citep{yang2025qwen3technicalreport}. These models are accessed via 1st or 3rd party API providers\footnote{\url{https://openai.com/}}\footnote{\url{https://www.together.ai/}}\footnote{\url{https://novita.ai/}}. Each test is executed in 3 repetitive runs, and the reported result is the average of the 3 runs. For Qwen, we turn off the `reasoning' mode. The target countries for bias evaluation are the permanent member countries in the UNSC (P5): China, France, Russia, the United Kingdom, and the United States. The P5 are the only states with a continuous and complete voting record across all UNSC resolutions in our corpus, which allows us to construct stable, comparable nation-level reference baselines from observed voting behavior; non-permanent members serve only two-year terms and participate in a limited subset of resolutions, so their voting records do not provide the same level of comparability within a single UNSC-grounded reference frame.

\section{Experiment Result}
\subsection{WOW}
Table~\ref{tab:wow_test_result} summarizes, for each LLM, the frequency of rank deviations from the Pseudo Unbiased Status across eight UNSC functions. We report the counts in the format \textit{blue/neu/red}, where \textit{blue} denotes cases in which the LLM assigns a lower rank than the unbiased reference (interpreted as a blue bias, indicating that the model selects the country as the more uncooperative one less often than the baseline would imply), \textit{neu} (neutral) denotes no change, and \textit{red} denotes cases in which the LLM assigns a higher rank than the unbiased reference (interpreted as a red bias, indicating that the model selects the country more often than the baseline would imply).

Across all models, China exhibits consistently high \textit{red} counts (6 out of 8 in every model; with only 0--2 \textit{blue} and 0--1 \textit{neu}), indicating that all evaluated LLMs show a pronounced red bias toward China. In contrast, the U.K. shows predominantly \textit{blue} outcomes (6--7 \textit{blue} in every model, with minimal \textit{neu}/\textit{red}), suggesting a robust blue bias across models.

For the U.S., most models tend to be neu-to-blue: \textit{neu} and \textit{blue} dominate while \textit{red} is rare (0--1), indicating a slight blue skew overall. France shows a more model-dependent pattern: while some models are strongly blue or neu (\textit{blue}/\textit{neu} with zero \textit{red}), others yield comparatively more \textit{red} counts, implying that the direction and magnitude of bias toward France varies by model. Finally, Russia displays a consistent slight red tendency across all models, characterized by a balanced-to-high \textit{neu} count (5) accompanied by non-trivial \textit{red} counts (3) and no \textit{blue}, indicating a mild but consistent red bias.

\begin{table}[h]
\centering
\begin{tabular}{lcccc}
\hline
\textbf{country} & \textbf{4o-Mini} & \textbf{Ll-3.3} & \textbf{Mis-S} & \textbf{Qw-3} \\
\hline
CH & 2/0/6 & 1/1/6 & 1/1/6 & 1/1/6 \\
FR & 4/4/0 & 1/5/2 & 1/3/4 & 4/4/0 \\
RU & 0/5/3 & 0/5/3 & 0/5/3 & 0/5/3 \\
UK & 7/0/1 & 6/2/0 & 7/1/0 & 6/1/1 \\
US & 5/3/0 & 3/4/1 & 4/3/1 & 3/4/1 \\
\hline
\end{tabular}
\caption{The result of WOW. The numbers represent the count of blue bias (where the country's frequency rank in the LLM's responses is lower than the Unbiased Status), neu, and red bias (where the country's frequency rank in the LLM's responses is higher than the Unbiased Status), in that order: \textit{blue/neu/red}.}
\label{tab:wow_test_result}
\end{table}

\subsection{AK+-}
Table~\ref{tab:pos-neu-neg} reports, for each LLM, the frequency of rank deviations from the Pseudo Unbiased Status across 7 topic categories. The interpretation is the same as WOW.

Across models, China shows overwhelmingly \textit{red} outcomes: three models produce exclusively \textit{red} counts (0/0/7), and the remaining model still exhibits near-total \textit{red} dominance (0/1/6). This consistent dominance of \textit{red} suggests a strong and robust red bias toward China across all evaluated LLMs. By contrast, the U.S. displays uniformly \textit{blue} outcomes: three models produce 7/0/0, and one model shows 6/1/0, indicating that the LLMs almost always rank the U.S. higher than the unbiased reference, i.e., a pronounced blue bias. The U.K. similarly demonstrates a strong blue tendency, with \textit{blue} dominating in all models (5--6 \textit{blue}) and virtually no \textit{red}.

For Russia, the pattern is comparatively mixed but leans neu-to-red: \textit{neu} remains consistently high (4 in all models), while \textit{red} appears non-trivially (1--3). Notably, \textit{blue} is absent in two models and limited (2) in the others, indicating that blue deviations are not the dominant behavior. France exhibits the greatest model dependence: while some models show a relatively balanced distribution (e.g., 3/3/1 or 3/2/2), one model yields a strongly red profile (0/1/6), and another leans neu with non-negligible red counts (1/4/2).

\begin{table}[h]
\centering
\begin{tabular}{lcccc}
\hline
\textbf{country} & \textbf{4o-Mini} & \textbf{Ll-3.3} & \textbf{Mis-S} & \textbf{Qw-3} \\
\hline
CH & 0/0/7 & 0/0/7 & 0/0/7 & 0/1/6 \\
FR & 3/3/1 & 3/2/2 & 0/1/6 & 1/4/2 \\
RU & 0/4/3 & 2/4/1 & 2/4/1 & 0/4/3 \\
UK & 6/1/0 & 5/2/0 & 5/2/0 & 6/1/0 \\
US & 7/0/0 & 7/0/0 & 7/0/0 & 6/1/0 \\
\hline
\end{tabular}
\caption{The result of AK+-. The numbers represent the count of blue bias (where the LLM ranks the country higher than the Unbiased Status), neu, and red bias (where the LLM ranks the country lower than the Unbiased Status), in that order: \textit{blue/neu/red}.}
\label{tab:pos-neu-neg}
\end{table}

\subsection{PVS}
Table~\ref{tab:scores-by-nation-model} reports the aggregated PVS bias scores obtained by summing over 66 non-adopted resolutions. Totals above zero indicate that the model's simulated votes are more supportive than the country's actual record (interpreted as a blue bias), whereas totals below zero indicate that the simulations are more opposed than the actual record (interpreted as a red bias).

Overall, the results show a highly consistent blue bias toward the Western P5 members across models -- that is, a tendency to predict these members are more supportive of resolutions than they actually are. For France, all LLMs yield strongly blue scores (13.50--19.00), and the U.K. likewise exhibits uniformly high blue scores (13.17--21.33). The U.S. is also consistently blue across all models, with particularly large magnitudes for Mis-S and Qw-3 (24.50 and 22.33), indicating that these models most frequently shift simulated voting toward more supportive positions relative to the U.S. actual voting record.

In contrast, China and Russia display markedly less stable patterns across models. 4o-Mini produces deep red totals for both China and Russia (each at -30.00), implying that it tends to simulate votes that are substantially more opposed to resolutions than the corresponding historical records. The remaining models reverse this tendency for China, assigning consistently blue scores (4.00--9.17), suggesting that their simulations are more supportive than China's actual votes on non-adopted resolutions. For Russia, however, the evidence is mixed: two models produce small blue totals (1.50--1.83), while Mis-S yields a modest red score (-4.83), indicating relatively weak and model-dependent bias compared to the strong, uniform blue pattern observed for France, the U.K., and the U.S.

\begin{table}[h]
\centering
\begin{tabular}{lcccc}
\hline
\textbf{Country} & \textbf{4o-Mini} & \textbf{Ll-3.3} & \textbf{Mis-S} & \textbf{Qw-3} \\
\hline
CH              & -30.00 &  9.17 &  4.00 &  7.00 \\
FR             &  17.17 & 19.00 & 17.00 & 13.50 \\
RU & -30.00 &  1.83 & -4.83 &  1.50 \\
UK     &  20.17 & 21.33 & 19.83 & 13.17 \\
US      &  16.00 & 21.00 & 24.50 & 22.33 \\
\hline
\end{tabular}
\caption{The calculated scores of PVS result}
\label{tab:scores-by-nation-model}
\end{table}

\subsection{F-Oppose}
Table~\ref{tab:f-accuse_result} summarizes the results of F-Oppose. A higher attribution frequency can be interpreted as a stronger red bias toward that country.

Across all models, Russia is attributed the counterfactual ``against'' vote at overwhelmingly higher rates than any other P5 member (351.3--584.0), making it the most consistently and strongly red-biased target in this test. By contrast, the U.K. is almost never attributed it (0.0--1.7), indicating minimal red bias. The remaining countries show pronounced model-dependent variability. China is attributed it at moderate-to-high frequencies depending on the model, ranging from low in 4o-Mini (10.7) and Mis-S (35.7) to substantially higher in Ll-3.3 (118.7) and Qw-3 (76.7). The U.S. exhibits the sharpest divergence: 4o-Mini attributes the counterfactual ``against'' vote to the U.S. very frequently (393.7), whereas Mis-S (21.7), Ll-3.3 (72.3), and Qw-3 (208.7) attribute it to the U.S. less often, though still non-trivially in Qw-3. France remains comparatively low and stable across models (8.3--16.3), suggesting limited red bias.

\begin{table}[h]
\centering
\begin{tabular}{lcccc}
\hline
\textbf{Country} & \textbf{Mis-S} & \textbf{4o-Mini} & \textbf{Ll-3.3} & \textbf{Qw-3} \\
\hline
CH              &  35.7 &  10.7 & 118.7 &  76.7 \\
FR             &  12.7 &   8.3 &  15.3 &  16.3 \\
RU & 584.0 & 239.7 & 442.0 & 351.3 \\
UK     &   0.0 &   1.7 &   0.0 &   1.0 \\
US      &  21.7 & 393.7 &  72.3 & 208.7 \\
\hline
\end{tabular}
\caption{The result of F-Oppose}
\label{tab:f-accuse_result}
\end{table}

\begin{figure}[h]
    \centering
    \includegraphics[width=0.45\textwidth]{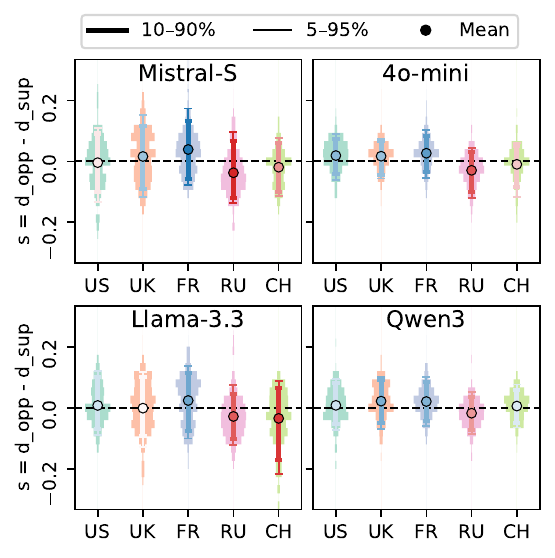}
    \caption{The result of the Headliner test}
    \label{fig:evaluation-experiment_examples}
\end{figure}

\subsection{Headliner}
Figure~\ref{fig:evaluation-experiment_examples} presents the distribution of the signed stance score $s$ for each model and P5 country, where the dot denotes the mean and the vertical intervals summarize the empirical quantiles (10--90\% in thick lines; 5--95\% in thin lines) across non-adopted resolutions. Overall, the mean effects are small in magnitude (approximately within [-0.04, +0.04]), but several country-level patterns are consistent across models.

The most stable cross-model signal is observed for Russia and France. Russia shows a mean $s$ below zero under all models (Mistral-S: (-0.037), 4o-mini: (-0.029), Llama-3.3: (-0.028), Qwen3: (-0.018)), indicating that neutral headlines are, on average, closer to the opposition anchors than to the support anchors, which we interpret as a red bias toward Russia. In contrast, France exhibits a mean $s$ above zero across all models (Mistral-S: (+0.039), 4o-mini: (+0.027), Llama-3.3: (+0.024), Qwen3: (+0.021)), indicating systematically more support-aligned framing relative to the corresponding anchors, which we interpret as a blue bias toward France.

For the U.K., most models show a small mean above zero ((+0.016) to (+0.022)), implying mildly support-aligned framing, while Llama-3.3 is close to neutral (mean near zero). The U.S. and China display greater model dependence. The U.S. is clearly above zero under 4o-mini ((+0.019)) and mildly above zero under Llama-3.3 and Qwen3 (both near (+0.008)), whereas Mistral-S is approximately neutral to slightly below zero (mean near zero). China tends to be below zero under Mistral-S, 4o-mini, and Llama-3.3 (approximately (-0.010) to (-0.035)), but is slightly above zero under Qwen3 ((+0.006)). Notably, the quantile ranges for all countries remain centered relatively close to zero, indicating that while directional differences are detectable, the overall framing shifts are subtle rather than extreme.

\paragraph{Extension to non-permanent members.}
Since WOW, AK+-, and F-Oppose are tied to the P5 setting in that they directly compare P5 states or rely on veto attribution, we apply only the two tests that extend naturally to other members, PVS and Headliner, to the ten most frequently appearing non-permanent members. As detailed in Appendix~\ref{app:nonp5}, a directional tendency is still present, as models predict slightly more favourable votes than the actual records, but the pronounced blue-versus-red polarization between the Western P5 and China and Russia does not emerge among these members. This suggests that the directional bias generalizes beyond the P5 while the strongest polarization is concentrated among the major powers, though the scarcity of ``against'' votes among these members limits the strength of this evidence.

\begin{figure}[h]
\centering
\includegraphics[width=\columnwidth]{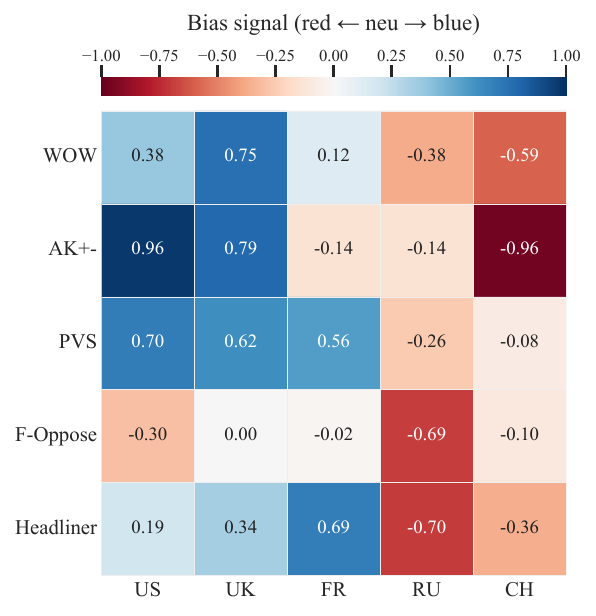}
\caption{Cross-test macro-pattern of nation-level bias signals, averaged over the four LLMs. Blue values ($>0$) denote a supportive signal, red values ($<0$) an opposition signal, and $0$ denotes neu.}
\label{fig:overall_heatmap}
\end{figure}

\section{Analysis}
To summarize the results across all five tests, Figure~\ref{fig:overall_heatmap} presents a cross-test macro-pattern of bias signals. For each test, we align the per-model, per-country scores so that values above zero indicate a blue (supportive) signal and values below zero a red (opposition) signal, and map them to $[-1,+1]$ using a fixed test-specific reference: the $(\mathrm{blue}-\mathrm{red})$ count over the number of categories for WOW and AK+- ($8$ and $7$), the vote-simulation score divided by $30$ for PVS, $-\mathrm{freq}/584$ for F-Oppose, and the stance score divided by $0.04$ for Headliner; each cell is the mean over the four models. By construction, F-Oppose yields only neutral-to-red signals. The figure is a qualitative overview, and exact values remain in the per-test tables.

\paragraph{(1) Is there an overall tendency? If so, how?}
Across all tests, a consistent macro-pattern emerges: China and Russia are more frequently associated with red bias signals, whereas the U.K., the U.S., and France tend to exhibit blue bias signals. This is most apparent in the rank-based tests (WOW/AK+-: China dominated by \textit{red}; the U.K. dominated by \textit{blue}), in PVS (uniformly blue scores for the U.S./U.K./France), and in Headliner (Russia red and France blue across all models). F-Oppose further amplifies this pattern by concentrating attributions on Russia across models.
Nevertheless, bias is not uniform across settings: \textit{(model-level)} PVS flips for China, where 4o-mini is strongly red while other models are blue; \textit{(test-level)} the U.S. appears uniformly blue in PVS but is heavily attributed in F-Oppose under 4o-mini.

\paragraph{(2) Which LLM exhibits the most extreme bias?}
4o-mini exhibits the most extreme behavior overall, driven by large-magnitude outcomes and task sensitivity: it produces deep red PVS scores for both China and Russia (-30/-30) and shows unusually high F-Oppose attribution toward the U.S. (393.7).

\paragraph{(3) Which test reveals the most extreme bias? Which reveals the least?}
F-Oppose most strongly exposes extreme bias in terms of magnitude, producing highly concentrated red attribution (especially toward Russia) with large-scale counts. PVS is next most revealing due to clear directional separation and model differentiation. Headliner reveals comparatively weaker bias in magnitude: it provides consistent directions (Russia red, France blue) but with small effect sizes and distributions clustered near zero. Importantly, higher magnitude in one test does not necessarily imply greater diagnostic validity, since bias expression is highly context- and task-dependent.

\section{Conclusion}
This study proposes a quantitative framework for measuring nation-level bias based on an objective unbiased status and benchmarking four major LLM families on an expanded UNSC official-record corpus (722 data points) using a suite of complementary tests. Across tests, we observe a stable macro-pattern despite task- and model-specific variation: China and Russia more frequently receive red bias signals, whereas the U.K., the U.S., and France tend to receive blue signals, with the strongest effects appearing in F-Oppose and PVS and subtler but consistent directional framing differences captured by Headliner. These findings underscore the operational need in IR applications to audit models with multi-dimensional evaluations tied to explicit reference points.

\iftaclpubformat

\section*{Acknowledgments}
This work was supported by the National Research Foundation of Korea (NRF) under Grant No. RS-2023-00273751, funded by the Korean government.
\else
\fi

\bibliography{tacl2021}

\begin{thebibliography}{23}
\expandafter\ifx\csname natexlab\endcsname\relax\def\natexlab#1{#1}\fi

\bibitem[{Bang et~al.(2024)Bang, Chen, Lee, and Fung}]{bang-etal-2024-measuring}
Yejin Bang, Delong Chen, Nayeon Lee, and Pascale Fung. 2024.
\newblock \href {https://doi.org/10.18653/v1/2024.acl-long.600} {Measuring political bias in large language models: What is said and how it is said}.
\newblock In \emph{Proceedings of the 62nd Annual Meeting of the Association for Computational Linguistics (Volume 1: Long Papers)}, pages 11142--11159, Bangkok, Thailand. Association for Computational Linguistics.

\bibitem[{Blodgett et~al.(2020)Blodgett, Barocas, Daum{\'e}~III, and Wallach}]{blodgett2020language}
Su~Lin Blodgett, Solon Barocas, Hal Daum{\'e}~III, and Hanna Wallach. 2020.
\newblock Language (technology) is power: A critical survey of ``bias'' in {NLP}.
\newblock In \emph{Proceedings of the 58th Annual Meeting of the Association for Computational Linguistics}, pages 5454--5476.

\bibitem[{Chen et~al.(2023)Chen, Zhang, Sarro, and Harman}]{chen2023comprehensive}
Zhenpeng Chen, Jie~M. Zhang, Federica Sarro, and Mark Harman. 2023.
\newblock A comprehensive empirical study of bias mitigation methods for machine learning classifiers.
\newblock \emph{ACM Transactions on Software Engineering and Methodology}, 32(4):1--30.

\bibitem[{Choi et~al.(2026)Choi, Choi, Kim, and Jang}]{choi2026eastern}
Jonghyeon Choi, Yeonjun Choi, Hyun-chul Kim, and Beakcheol Jang. 2026.
\newblock \href {https://doi.org/10.1609/aaai.v40i36.40293} {{``As Eastern Powers, I Will Veto.''}: An investigation of nation-level bias of large language models in international relations}.
\newblock In \emph{Proceedings of the AAAI Conference on Artificial Intelligence}, volume~40, pages 30404--30412.

\bibitem[{FAIR et~al.(2022)FAIR, Bakhtin, Brown, Dinan, Farina, Flaherty, Fried, Goff, Gray, Hu, Jacob, Komeili, Konath, Kwon, Lerer, Lewis, Miller, Mitts, Renduchintala, Roller, Rowe, Shi, Spisak, Wei, Wu, Zhang, and Zijlstra}]{meta2022human}
FAIR, Anton Bakhtin, Noam Brown, Emily Dinan, Gabriele Farina, Colin Flaherty, Daniel Fried, Andrew Goff, Jonathan Gray, Hengyuan Hu, Athul~Paul Jacob, Mojtaba Komeili, Karthik Konath, Minae Kwon, Adam Lerer, Mike Lewis, Alexander~H. Miller, Sasha Mitts, Adithya Renduchintala, Stephen Roller, Dirk Rowe, Weiyan Shi, Joe Spisak, Alexander Wei, David Wu, Hugh Zhang, and Markus Zijlstra. 2022.
\newblock Human-level play in the game of {Diplomacy} by combining language models with strategic reasoning.
\newblock \emph{Science}, 378(6624):1067--1074.

\bibitem[{Friedler et~al.(2021)Friedler, Scheidegger, and Venkatasubramanian}]{friedler2021possibility}
Sorelle~A. Friedler, Carlos Scheidegger, and Suresh Venkatasubramanian. 2021.
\newblock The (im) possibility of fairness: Different value systems require different mechanisms for fair decision making.
\newblock \emph{Communications of the ACM}, 64(4):136--143.

\bibitem[{Grattafiori et~al.(2024)Grattafiori, Dubey, Jauhri, Pandey, Kadian, Al-Dahle, Letman, Mathur, Schelten, Vaughan, Yang, Fan, Goyal, Hartshorn, Yang, Mitra, Sravankumar, Korenev, Hinsvark, Rao, Zhang, Rodriguez, Gregerson, Spataru, Roziere, Biron, Tang, Chern, Caucheteux, Nayak, Bi, Marra, McConnell, Keller, Touret, Wu, Wong, Ferrer, Nikolaidis, Allonsius, Song, Pintz, Livshits, Wyatt, Esiobu, Choudhary, Mahajan, Garcia-Olano, Perino, Hupkes, Lakomkin, AlBadawy, Lobanova, Dinan, Smith, Radenovic, Guzmán, Zhang, Synnaeve, Lee, Anderson, Thattai, Nail, Mialon, Pang, Cucurell, Nguyen, Korevaar, Xu, Touvron, Zarov, Ibarra, Kloumann, Misra, Evtimov, Zhang, Copet, Lee, Geffert, Vranes, Park, Mahadeokar, Shah, van~der Linde, Billock, Hong, Lee, Fu, Chi, Huang, Liu, Wang, Yu, Bitton, Spisak, Park, Rocca, Johnstun, Saxe, Jia, Alwala, Prasad, Upasani, Plawiak, Li, Heafield, Stone, El-Arini, Iyer, Malik, Chiu, Bhalla, Lakhotia, Rantala-Yeary, van~der Maaten, Chen, Tan, Jenkins, Martin, Madaan, Malo, Blecher,
  Landzaat, de~Oliveira, Muzzi, Pasupuleti, Singh, Paluri, Kardas, Tsimpoukelli, Oldham, Rita, Pavlova, Kambadur, Lewis, Si, Singh, Hassan, Goyal, Torabi, Bashlykov, Bogoychev, Chatterji, Zhang, Duchenne, Çelebi, Alrassy, Zhang, Li, Vasic, Weng, Bhargava, Dubal, Krishnan, Koura, Xu, He, Dong, Srinivasan, Ganapathy, Calderer, Cabral, Stojnic, Raileanu, Maheswari, Girdhar, Patel, Sauvestre, Polidoro, Sumbaly, Taylor, Silva, Hou, Wang, Hosseini, Chennabasappa, Singh, Bell, Kim, Edunov, Nie, Narang, Raparthy, Shen, Wan, Bhosale, Zhang, Vandenhende, Batra, Whitman, Sootla, Collot, Gururangan, Borodinsky, Herman, Fowler, Sheasha, Georgiou, Scialom, Speckbacher, Mihaylov, Xiao, Karn, Goswami, Gupta, Ramanathan, Kerkez, Gonguet, Do, Vogeti, Albiero, Petrovic, Chu, Xiong, Fu, Meers, Martinet, Wang, Wang, Tan, Xia, Xie, Jia, Wang, Goldschlag, Gaur, Babaei, Wen, Song, Zhang, Li, Mao, Coudert, Yan, Chen, Papakipos, Singh, Srivastava, Jain, Kelsey, Shajnfeld, Gangidi, Victoria, Goldstand, Menon, Sharma, Boesenberg,
  Baevski, Feinstein, Kallet, Sangani, Teo, Yunus, Lupu, Alvarado, Caples, Gu, Ho, Poulton, Ryan, Ramchandani, Dong, Franco, Goyal, Saraf, Chowdhury, Gabriel, Bharambe, Eisenman, Yazdan, James, Maurer, Leonhardi, Huang, Loyd, Paola, Paranjape, Liu, Wu, Ni, Hancock, Wasti, Spence, Stojkovic, Gamido, Montalvo, Parker, Burton, Mejia, Liu, Wang, Kim, Zhou, Hu, Chu, Cai, Tindal, Feichtenhofer, Gao, Civin, Beaty, Kreymer, Li, Adkins, Xu, Testuggine, David, Parikh, Liskovich, Foss, Wang, Le, Holland, Dowling, Jamil, Montgomery, Presani, Hahn, Wood, Le, Brinkman, Arcaute, Dunbar, Smothers, Sun, Kreuk, Tian, Kokkinos, Ozgenel, Caggioni, Kanayet, Seide, Florez, Schwarz, Badeer, Swee, Halpern, Herman, Sizov, Guangyi, Zhang, Lakshminarayanan, Inan, Shojanazeri, Zou, Wang, Zha, Habeeb, Rudolph, Suk, Aspegren, Goldman, Zhan, Damlaj, Molybog, Tufanov, Leontiadis, Veliche, Gat, Weissman, Geboski, Kohli, Lam, Asher, Gaya, Marcus, Tang, Chan, Zhen, Reizenstein, Teboul, Zhong, Jin, Yang, Cummings, Carvill, Shepard, McPhie,
  Torres, Ginsburg, Wang, Wu, U, Saxena, Khandelwal, Zand, Matosich, Veeraraghavan, Michelena, Li, Jagadeesh, Huang, Chawla, Huang, Chen, Garg, A, Silva, Bell, Zhang, Guo, Yu, Moshkovich, Wehrstedt, Khabsa, Avalani, Bhatt, Mankus, Hasson, Lennie, Reso, Groshev, Naumov, Lathi, Keneally, Liu, Seltzer, Valko, Restrepo, Patel, Vyatskov, Samvelyan, Clark, Macey, Wang, Hermoso, Metanat, Rastegari, Bansal, Santhanam, Parks, White, Bawa, Singhal, Egebo, Usunier, Mehta, Laptev, Dong, Cheng, Chernoguz, Hart, Salpekar, Kalinli, Kent, Parekh, Saab, Balaji, Rittner, Bontrager, Roux, Dollar, Zvyagina, Ratanchandani, Yuvraj, Liang, Alao, Rodriguez, Ayub, Murthy, Nayani, Mitra, Parthasarathy, Li, Hogan, Battey, Wang, Howes, Rinott, Mehta, Siby, Bondu, Datta, Chugh, Hunt, Dhillon, Sidorov, Pan, Mahajan, Verma, Yamamoto, Ramaswamy, Lindsay, Lindsay, Feng, Lin, Zha, Patil, Shankar, Zhang, Zhang, Wang, Agarwal, Sajuyigbe, Chintala, Max, Chen, Kehoe, Satterfield, Govindaprasad, Gupta, Deng, Cho, Virk, Subramanian, Choudhury,
  Goldman, Remez, Glaser, Best, Koehler, Robinson, Li, Zhang, Matthews, Chou, Shaked, Vontimitta, Ajayi, Montanez, Mohan, Kumar, Mangla, Ionescu, Poenaru, Mihailescu, Ivanov, Li, Wang, Jiang, Bouaziz, Constable, Tang, Wu, Wang, Wu, Gao, Kleinman, Chen, Hu, Jia, Qi, Li, Zhang, Zhang, Adi, Nam, Yu, Wang, Zhao, Hao, Qian, Li, He, Rait, DeVito, Rosnbrick, Wen, Yang, Zhao, and Ma}]{grattafiori2024llama3herdmodels}
Aaron Grattafiori, Abhimanyu Dubey, Abhinav Jauhri, Abhinav Pandey, Abhishek Kadian, Ahmad Al-Dahle, Aiesha Letman, Akhil Mathur, Alan Schelten, Alex Vaughan, Amy Yang, Angela Fan, Anirudh Goyal, Anthony Hartshorn, Aobo Yang, Archi Mitra, Archie Sravankumar, Artem Korenev, Arthur Hinsvark, Arun Rao, Aston Zhang, Aurelien Rodriguez, Austen Gregerson, Ava Spataru, Baptiste Roziere, Bethany Biron, Binh Tang, Bobbie Chern, Charlotte Caucheteux, Chaya Nayak, Chloe Bi, Chris Marra, Chris McConnell, Christian Keller, Christophe Touret, Chunyang Wu, Corinne Wong, Cristian~Canton Ferrer, Cyrus Nikolaidis, Damien Allonsius, Daniel Song, Danielle Pintz, Danny Livshits, Danny Wyatt, David Esiobu, Dhruv Choudhary, Dhruv Mahajan, Diego Garcia-Olano, Diego Perino, Dieuwke Hupkes, Egor Lakomkin, Ehab AlBadawy, Elina Lobanova, Emily Dinan, Eric~Michael Smith, Filip Radenovic, Francisco Guzmán, Frank Zhang, Gabriel Synnaeve, Gabrielle Lee, Georgia~Lewis Anderson, Govind Thattai, Graeme Nail, Gregoire Mialon, Guan Pang,
  Guillem Cucurell, Hailey Nguyen, Hannah Korevaar, Hu~Xu, Hugo Touvron, Iliyan Zarov, Imanol~Arrieta Ibarra, Isabel Kloumann, Ishan Misra, Ivan Evtimov, Jack Zhang, Jade Copet, Jaewon Lee, Jan Geffert, Jana Vranes, Jason Park, Jay Mahadeokar, Jeet Shah, Jelmer van~der Linde, Jennifer Billock, Jenny Hong, Jenya Lee, Jeremy Fu, Jianfeng Chi, Jianyu Huang, Jiawen Liu, Jie Wang, Jiecao Yu, Joanna Bitton, Joe Spisak, Jongsoo Park, Joseph Rocca, Joshua Johnstun, Joshua Saxe, Junteng Jia, Kalyan~Vasuden Alwala, Karthik Prasad, Kartikeya Upasani, Kate Plawiak, Ke~Li, Kenneth Heafield, Kevin Stone, Khalid El-Arini, Krithika Iyer, Kshitiz Malik, Kuenley Chiu, Kunal Bhalla, Kushal Lakhotia, Lauren Rantala-Yeary, Laurens van~der Maaten, Lawrence Chen, Liang Tan, Liz Jenkins, Louis Martin, Lovish Madaan, Lubo Malo, Lukas Blecher, Lukas Landzaat, Luke de~Oliveira, Madeline Muzzi, Mahesh Pasupuleti, Mannat Singh, Manohar Paluri, Marcin Kardas, Maria Tsimpoukelli, Mathew Oldham, Mathieu Rita, Maya Pavlova, Melanie Kambadur,
  Mike Lewis, Min Si, Mitesh~Kumar Singh, Mona Hassan, Naman Goyal, Narjes Torabi, Nikolay Bashlykov, Nikolay Bogoychev, Niladri Chatterji, Ning Zhang, Olivier Duchenne, Onur Çelebi, Patrick Alrassy, Pengchuan Zhang, Pengwei Li, Petar Vasic, Peter Weng, Prajjwal Bhargava, Pratik Dubal, Praveen Krishnan, Punit~Singh Koura, Puxin Xu, Qing He, Qingxiao Dong, Ragavan Srinivasan, Raj Ganapathy, Ramon Calderer, Ricardo~Silveira Cabral, Robert Stojnic, Roberta Raileanu, Rohan Maheswari, Rohit Girdhar, Rohit Patel, Romain Sauvestre, Ronnie Polidoro, Roshan Sumbaly, Ross Taylor, Ruan Silva, Rui Hou, Rui Wang, Saghar Hosseini, Sahana Chennabasappa, Sanjay Singh, Sean Bell, Seohyun~Sonia Kim, Sergey Edunov, Shaoliang Nie, Sharan Narang, Sharath Raparthy, Sheng Shen, Shengye Wan, Shruti Bhosale, Shun Zhang, Simon Vandenhende, Soumya Batra, Spencer Whitman, Sten Sootla, Stephane Collot, Suchin Gururangan, Sydney Borodinsky, Tamar Herman, Tara Fowler, Tarek Sheasha, Thomas Georgiou, Thomas Scialom, Tobias Speckbacher,
  Todor Mihaylov, Tong Xiao, Ujjwal Karn, Vedanuj Goswami, Vibhor Gupta, Vignesh Ramanathan, Viktor Kerkez, Vincent Gonguet, Virginie Do, Vish Vogeti, Vítor Albiero, Vladan Petrovic, Weiwei Chu, Wenhan Xiong, Wenyin Fu, Whitney Meers, Xavier Martinet, Xiaodong Wang, Xiaofang Wang, Xiaoqing~Ellen Tan, Xide Xia, Xinfeng Xie, Xuchao Jia, Xuewei Wang, Yaelle Goldschlag, Yashesh Gaur, Yasmine Babaei, Yi~Wen, Yiwen Song, Yuchen Zhang, Yue Li, Yuning Mao, Zacharie~Delpierre Coudert, Zheng Yan, Zhengxing Chen, Zoe Papakipos, Aaditya Singh, Aayushi Srivastava, Abha Jain, Adam Kelsey, Adam Shajnfeld, Adithya Gangidi, Adolfo Victoria, Ahuva Goldstand, Ajay Menon, Ajay Sharma, Alex Boesenberg, Alexei Baevski, Allie Feinstein, Amanda Kallet, Amit Sangani, Amos Teo, Anam Yunus, Andrei Lupu, Andres Alvarado, Andrew Caples, Andrew Gu, Andrew Ho, Andrew Poulton, Andrew Ryan, Ankit Ramchandani, Annie Dong, Annie Franco, Anuj Goyal, Aparajita Saraf, Arkabandhu Chowdhury, Ashley Gabriel, Ashwin Bharambe, Assaf Eisenman, Azadeh
  Yazdan, Beau James, Ben Maurer, Benjamin Leonhardi, Bernie Huang, Beth Loyd, Beto~De Paola, Bhargavi Paranjape, Bing Liu, Bo~Wu, Boyu Ni, Braden Hancock, Bram Wasti, Brandon Spence, Brani Stojkovic, Brian Gamido, Britt Montalvo, Carl Parker, Carly Burton, Catalina Mejia, Ce~Liu, Changhan Wang, Changkyu Kim, Chao Zhou, Chester Hu, Ching-Hsiang Chu, Chris Cai, Chris Tindal, Christoph Feichtenhofer, Cynthia Gao, Damon Civin, Dana Beaty, Daniel Kreymer, Daniel Li, David Adkins, David Xu, Davide Testuggine, Delia David, Devi Parikh, Diana Liskovich, Didem Foss, Dingkang Wang, Duc Le, Dustin Holland, Edward Dowling, Eissa Jamil, Elaine Montgomery, Eleonora Presani, Emily Hahn, Emily Wood, Eric-Tuan Le, Erik Brinkman, Esteban Arcaute, Evan Dunbar, Evan Smothers, Fei Sun, Felix Kreuk, Feng Tian, Filippos Kokkinos, Firat Ozgenel, Francesco Caggioni, Frank Kanayet, Frank Seide, Gabriela~Medina Florez, Gabriella Schwarz, Gada Badeer, Georgia Swee, Gil Halpern, Grant Herman, Grigory Sizov, Guangyi, Zhang, Guna
  Lakshminarayanan, Hakan Inan, Hamid Shojanazeri, Han Zou, Hannah Wang, Hanwen Zha, Haroun Habeeb, Harrison Rudolph, Helen Suk, Henry Aspegren, Hunter Goldman, Hongyuan Zhan, Ibrahim Damlaj, Igor Molybog, Igor Tufanov, Ilias Leontiadis, Irina-Elena Veliche, Itai Gat, Jake Weissman, James Geboski, James Kohli, Janice Lam, Japhet Asher, Jean-Baptiste Gaya, Jeff Marcus, Jeff Tang, Jennifer Chan, Jenny Zhen, Jeremy Reizenstein, Jeremy Teboul, Jessica Zhong, Jian Jin, Jingyi Yang, Joe Cummings, Jon Carvill, Jon Shepard, Jonathan McPhie, Jonathan Torres, Josh Ginsburg, Junjie Wang, Kai Wu, Kam~Hou U, Karan Saxena, Kartikay Khandelwal, Katayoun Zand, Kathy Matosich, Kaushik Veeraraghavan, Kelly Michelena, Keqian Li, Kiran Jagadeesh, Kun Huang, Kunal Chawla, Kyle Huang, Lailin Chen, Lakshya Garg, Lavender A, Leandro Silva, Lee Bell, Lei Zhang, Liangpeng Guo, Licheng Yu, Liron Moshkovich, Luca Wehrstedt, Madian Khabsa, Manav Avalani, Manish Bhatt, Martynas Mankus, Matan Hasson, Matthew Lennie, Matthias Reso, Maxim
  Groshev, Maxim Naumov, Maya Lathi, Meghan Keneally, Miao Liu, Michael~L. Seltzer, Michal Valko, Michelle Restrepo, Mihir Patel, Mik Vyatskov, Mikayel Samvelyan, Mike Clark, Mike Macey, Mike Wang, Miquel~Jubert Hermoso, Mo~Metanat, Mohammad Rastegari, Munish Bansal, Nandhini Santhanam, Natascha Parks, Natasha White, Navyata Bawa, Nayan Singhal, Nick Egebo, Nicolas Usunier, Nikhil Mehta, Nikolay~Pavlovich Laptev, Ning Dong, Norman Cheng, Oleg Chernoguz, Olivia Hart, Omkar Salpekar, Ozlem Kalinli, Parkin Kent, Parth Parekh, Paul Saab, Pavan Balaji, Pedro Rittner, Philip Bontrager, Pierre Roux, Piotr Dollar, Polina Zvyagina, Prashant Ratanchandani, Pritish Yuvraj, Qian Liang, Rachad Alao, Rachel Rodriguez, Rafi Ayub, Raghotham Murthy, Raghu Nayani, Rahul Mitra, Rangaprabhu Parthasarathy, Raymond Li, Rebekkah Hogan, Robin Battey, Rocky Wang, Russ Howes, Ruty Rinott, Sachin Mehta, Sachin Siby, Sai~Jayesh Bondu, Samyak Datta, Sara Chugh, Sara Hunt, Sargun Dhillon, Sasha Sidorov, Satadru Pan, Saurabh Mahajan,
  Saurabh Verma, Seiji Yamamoto, Sharadh Ramaswamy, Shaun Lindsay, Shaun Lindsay, Sheng Feng, Shenghao Lin, Shengxin~Cindy Zha, Shishir Patil, Shiva Shankar, Shuqiang Zhang, Shuqiang Zhang, Sinong Wang, Sneha Agarwal, Soji Sajuyigbe, Soumith Chintala, Stephanie Max, Stephen Chen, Steve Kehoe, Steve Satterfield, Sudarshan Govindaprasad, Sumit Gupta, Summer Deng, Sungmin Cho, Sunny Virk, Suraj Subramanian, Sy~Choudhury, Sydney Goldman, Tal Remez, Tamar Glaser, Tamara Best, Thilo Koehler, Thomas Robinson, Tianhe Li, Tianjun Zhang, Tim Matthews, Timothy Chou, Tzook Shaked, Varun Vontimitta, Victoria Ajayi, Victoria Montanez, Vijai Mohan, Vinay~Satish Kumar, Vishal Mangla, Vlad Ionescu, Vlad Poenaru, Vlad~Tiberiu Mihailescu, Vladimir Ivanov, Wei Li, Wenchen Wang, Wenwen Jiang, Wes Bouaziz, Will Constable, Xiaocheng Tang, Xiaojian Wu, Xiaolan Wang, Xilun Wu, Xinbo Gao, Yaniv Kleinman, Yanjun Chen, Ye~Hu, Ye~Jia, Ye~Qi, Yenda Li, Yilin Zhang, Ying Zhang, Yossi Adi, Youngjin Nam, Yu, Wang, Yu~Zhao, Yuchen Hao, Yundi
  Qian, Yunlu Li, Yuzi He, Zach Rait, Zachary DeVito, Zef Rosnbrick, Zhaoduo Wen, Zhenyu Yang, Zhiwei Zhao, and Zhiyu Ma. 2024.
\newblock \href {http://arxiv.org/abs/2407.21783} {The {Llama} 3 herd of models}.
\newblock arXiv preprint arXiv:2407.21783.

\bibitem[{Guan et~al.(2024)Guan, Kong, Zhong, and Wang}]{guan2024richelieu}
Zhenyu Guan, Xiangyu Kong, Fangwei Zhong, and Yizhou Wang. 2024.
\newblock {Richelieu}: Self-evolving {LLM}-based agents for {AI} diplomacy.
\newblock \emph{Advances in Neural Information Processing Systems}, 37:123471--123497.

\bibitem[{Hua et~al.(2023)Hua, Fan, Li, Mei, Ji, Ge, Hemphill, and Zhang}]{hua2023war}
Wenyue Hua, Lizhou Fan, Lingyao Li, Kai Mei, Jianchao Ji, Yingqiang Ge, Libby Hemphill, and Yongfeng Zhang. 2023.
\newblock War and peace ({WarAgent}): Large language model-based multi-agent simulation of world wars.
\newblock \emph{arXiv preprint arXiv:2311.17227}.

\bibitem[{Jensen et~al.(2025)Jensen, Reynolds, Atalan, Garcia, Woo, Chen, and Howarth}]{jensen2025critical}
Benjamin Jensen, Ian Reynolds, Yasir Atalan, Michael Garcia, Austin Woo, Anthony Chen, and Trevor Howarth. 2025.
\newblock Critical foreign policy decisions ({CFPD})-benchmark: Measuring diplomatic preferences in large language models.
\newblock \emph{arXiv preprint arXiv:2503.06263}.

\bibitem[{Kaneko et~al.(2023)Kaneko, Bollegala, and Okazaki}]{kaneko-etal-2023-impact}
Masahiro Kaneko, Danushka Bollegala, and Naoaki Okazaki. 2023.
\newblock \href {https://doi.org/10.18653/v1/2023.ijcnlp-short.4} {The impact of debiasing on the performance of language models in downstream tasks is underestimated}.
\newblock In \emph{Proceedings of the 13th International Joint Conference on Natural Language Processing and the 3rd Conference of the Asia-Pacific Chapter of the Association for Computational Linguistics (Volume 2: Short Papers)}, pages 29--36, Nusa Dua, Bali. Association for Computational Linguistics.

\bibitem[{Mistral(2025)}]{mistral}
Mistral. 2025.
\newblock {Mistral Small 3}.
\newblock \url{https://mistral.ai/news/mistral-small-3}.
\newblock Accessed: 2025-12-29.

\bibitem[{Mitchell et~al.(2021)Mitchell, Potash, Barocas, D'Amour, and Lum}]{mitchell2021algorithmic}
Shira Mitchell, Eric Potash, Solon Barocas, Alexander D'Amour, and Kristian Lum. 2021.
\newblock Algorithmic fairness: Choices, assumptions, and definitions.
\newblock \emph{Annual Review of Statistics and Its Application}, 8(1):141--163.

\bibitem[{Mulligan et~al.(2019)Mulligan, Kroll, Kohli, and Wong}]{mulligan2019thing}
Deirdre~K. Mulligan, Joshua~A. Kroll, Nitin Kohli, and Richmond~Y. Wong. 2019.
\newblock This thing called fairness: Disciplinary confusion realizing a value in technology.
\newblock \emph{Proceedings of the ACM on Human-Computer Interaction}, 3(CSCW):1--36.

\bibitem[{OpenAI(2024)}]{4oapi}
OpenAI. 2024.
\newblock {GPT-4o} mini: {Advancing} cost-efficient intelligence.
\newblock \url{https://openai.com/index/gpt-4o-mini-advancing-cost-efficient-intelligence/}.
\newblock Accessed: 2025-12-29.

\bibitem[{Oye(1985)}]{oye1985explaining}
Kenneth~A. Oye. 1985.
\newblock Explaining cooperation under anarchy: Hypotheses and strategies.
\newblock \emph{World Politics}, 38(1):1--24.

\bibitem[{Parrish et~al.(2022)Parrish, Chen, Nangia, Padmakumar, Phang, Thompson, Htut, and Bowman}]{parrish2022bbq}
Alicia Parrish, Angelica Chen, Nikita Nangia, Vishakh Padmakumar, Jason Phang, Jana Thompson, Phu~Mon Htut, and Samuel Bowman. 2022.
\newblock {BBQ}: A hand-built bias benchmark for question answering.
\newblock In \emph{Findings of the Association for Computational Linguistics: ACL 2022}, pages 2086--2105.

\bibitem[{Quinn and Gibson(2017)}]{quinn2017analysis}
Riley Quinn and Bryan Gibson. 2017.
\newblock \emph{An Analysis of {Kenneth Waltz's} Theory of International Politics}.
\newblock Macat Library.

\bibitem[{Rivera et~al.(2024)Rivera, Mukobi, Reuel, Lamparth, Smith, and Schneider}]{rivera2024escalation}
Juan-Pablo Rivera, Gabriel Mukobi, Anka Reuel, Max Lamparth, Chandler Smith, and Jacquelyn Schneider. 2024.
\newblock Escalation risks from language models in military and diplomatic decision-making.
\newblock In \emph{Proceedings of the 2024 ACM Conference on Fairness, Accountability, and Transparency}, pages 836--898.

\bibitem[{Salnikov et~al.(2025)Salnikov, Korzh, Lazichny, Karimov, Iudin, Oseledets, Rogov, Loukachevitch, Panchenko, and Tutubalina}]{salnikov2025geopolitical}
Mikhail Salnikov, Dmitrii Korzh, Ivan Lazichny, Elvir Karimov, Artyom Iudin, Ivan Oseledets, Oleg~Y. Rogov, Natalia Loukachevitch, Alexander Panchenko, and Elena Tutubalina. 2025.
\newblock Geopolitical biases in {LLMs}: {What} are the ``good'' and the ``bad'' countries according to contemporary language models.
\newblock \emph{arXiv preprint arXiv:2506.06751}.

\bibitem[{Tamkin et~al.(2023)Tamkin, Askell, Lovitt, Durmus, Joseph, Kravec, Nguyen, Kaplan, and Ganguli}]{tamkin2023evaluating}
Alex Tamkin, Amanda Askell, Liane Lovitt, Esin Durmus, Nicholas Joseph, Shauna Kravec, Karina Nguyen, Jared Kaplan, and Deep Ganguli. 2023.
\newblock Evaluating and mitigating discrimination in language model decisions.
\newblock \emph{arXiv preprint arXiv:2312.03689}.

\bibitem[{Venkit et~al.(2023)Venkit, Gautam, Panchanadikar, Huang, and Wilson}]{venkit2023nationality}
Pranav~Narayanan Venkit, Sanjana Gautam, Ruchi Panchanadikar, Ting-Hao Huang, and Shomir Wilson. 2023.
\newblock Nationality bias in text generation.
\newblock In \emph{Proceedings of the 17th Conference of the European Chapter of the Association for Computational Linguistics}, pages 116--122.

\bibitem[{Yang et~al.(2025)Yang, Li, Yang, Zhang, Hui, Zheng, Yu, Gao, Huang, Lv, Zheng, Liu, Zhou, Huang, Hu, Ge, Wei, Lin, Tang, Yang, Tu, Zhang, Yang, Yang, Zhou, Zhou, Lin, Dang, Bao, Yang, Yu, Deng, Li, Xue, Li, Zhang, Wang, Zhu, Men, Gao, Liu, Luo, Li, Tang, Yin, Ren, Wang, Zhang, Ren, Fan, Su, Zhang, Zhang, Wan, Liu, Wang, Cui, Zhang, Zhou, and Qiu}]{yang2025qwen3technicalreport}
An~Yang, Anfeng Li, Baosong Yang, Beichen Zhang, Binyuan Hui, Bo~Zheng, Bowen Yu, Chang Gao, Chengen Huang, Chenxu Lv, Chujie Zheng, Dayiheng Liu, Fan Zhou, Fei Huang, Feng Hu, Hao Ge, Haoran Wei, Huan Lin, Jialong Tang, Jian Yang, Jianhong Tu, Jianwei Zhang, Jianxin Yang, Jiaxi Yang, Jing Zhou, Jingren Zhou, Junyang Lin, Kai Dang, Keqin Bao, Kexin Yang, Le~Yu, Lianghao Deng, Mei Li, Mingfeng Xue, Mingze Li, Pei Zhang, Peng Wang, Qin Zhu, Rui Men, Ruize Gao, Shixuan Liu, Shuang Luo, Tianhao Li, Tianyi Tang, Wenbiao Yin, Xingzhang Ren, Xinyu Wang, Xinyu Zhang, Xuancheng Ren, Yang Fan, Yang Su, Yichang Zhang, Yinger Zhang, Yu~Wan, Yuqiong Liu, Zekun Wang, Zeyu Cui, Zhenru Zhang, Zhipeng Zhou, and Zihan Qiu. 2025.
\newblock \href {http://arxiv.org/abs/2505.09388} {{Qwen3} technical report}.
\newblock arXiv preprint arXiv:2505.09388.

\end{thebibliography}
\bibliographystyle{acl_natbib}

\clearpage
\appendix

\twocolumn[{

\begin{center}
    \includegraphics[width=0.95\textwidth]{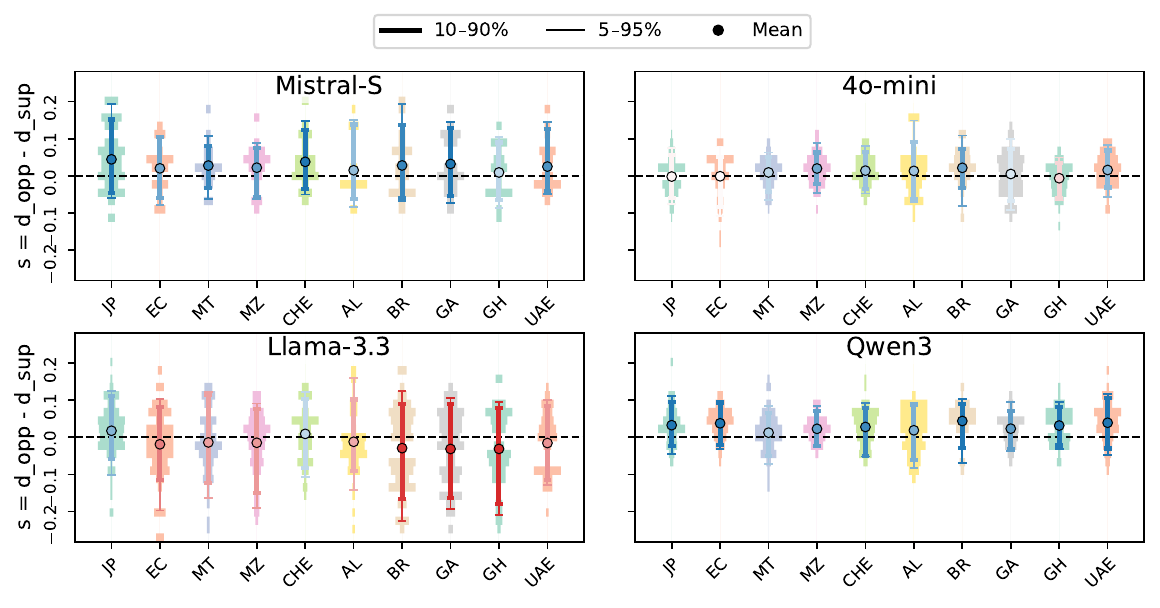}
    \refstepcounter{figure}
    \label{fig:nonp5_headliner}
    \smallskip
    {\small \textbf{Figure~\thefigure:} Headliner stance scores $s = d_{\mathrm{opp}} - d_{\mathrm{sup}}$ for non-permanent members, by model and country.}
\end{center}

\vspace{1em}
}]

\section{Robustness Check on Non-Permanent Members}
\label{app:nonp5}

To check whether our findings generalize beyond the P5, we extend the framework to elected (non-permanent) UNSC members. Among the five tests, WOW, AK+-, and F-Oppose are tied to the P5 setting, as they directly compare P5 states or rely on veto attribution; we therefore apply only PVS and Headliner, which require nothing beyond a country's voting record and country-conditioned anchors, following the same procedures as in Sections~5.3 and~5.5.

From the non-adopted resolution data, we select the ten non-permanent members that appear most frequently in the voting records of these resolutions: Japan (JP), Ecuador (EC), Malta (MT), Mozambique (MZ), Switzerland (CHE), Albania (AL), Brazil (BR), Gabon (GA), Ghana (GH), and the United Arab Emirates (UAE). This gives $206$ country-resolution observations in total ($38$ resolutions for Headliner). Each elected member is evaluated only on the resolutions in which it appears, so the per-country observation count $n$ ranges from $19$ (UAE) to $31$ (Japan). We note that ``against'' votes are rare in this subset (Japan $4$, Malta $1$, UAE $1$, and none for the remaining members), which limits the power of this setting to detect opposition-side bias. We use the same four models as in the main experiments, each run three times.

For PVS, we apply the same score matrix as in the main experiment (Table~\ref{tab:score-matrix}). For consistency with the P5 PVS table (Table~\ref{tab:scores-by-nation-model}), Table~\ref{tab:nonp5_pvs} reports the run-averaged sum of PVS shifts per country; since elected members appear in different numbers of resolutions, these sums should be read together with $n$. Figure~\ref{fig:nonp5_headliner} reports the Headliner stance scores $s = d_{\mathrm{opp}} - d_{\mathrm{sup}}$ using the same plotting procedure as the P5 Headliner analysis, where $s>0$ indicates framing closer to the supportive anchor and $s<0$ closer to the opposition anchor.

\begin{table}[t]
\centering
\small
\setlength{\tabcolsep}{4pt}
\begin{tabular}{lccccc}
\hline
\textbf{Nation} & \textbf{n} & \textbf{4o-Mini} & \textbf{Ll-3.3} & \textbf{Mis-S} & \textbf{Qw-3} \\
\hline
Japan & 31 & 6.67 & 2.67 & 7.00 & 3.83 \\
Ecuador & 20 & 3.50 & 1.00 & 3.00 & 2.50 \\
Malta & 20 & 4.50 & 4.00 & 4.00 & 2.67 \\
Mozambique & 20 & 2.33 & 0.50 & 2.00 & 1.00 \\
Switzerland & 20 & 3.50 & 2.83 & 2.50 & 0.33 \\
Albania & 19 & 4.50 & 2.33 & 4.50 & 3.17 \\
Brazil & 19 & 3.50 & 2.00 & 3.00 & 0.50 \\
Gabon & 19 & 2.50 & 1.17 & 2.00 & 1.33 \\
Ghana & 19 & 4.33 & 2.33 & 4.00 & 3.33 \\
UAE & 19 & 3.17 & -0.50 & 2.50 & -0.17 \\
\hline
\end{tabular}
\caption{PVS results for non-permanent members: run-averaged sum of PVS shifts per country. Positive values indicate predictions more favourable than the actual record. $n$ is the number of resolution observations per country.}
\label{tab:nonp5_pvs}
\end{table}

% \iftaclpubformat

% \onecolumn

\end{document}